\pdfoutput=1
\documentclass[11pt]{article}

\usepackage[]{./template/acl}

\usepackage{times}
\usepackage{latexsym}

\usepackage[T1]{fontenc}

\usepackage[utf8]{inputenc}

\usepackage{microtype}

\usepackage{soul}
\usepackage[ruled]{algorithm2e}
\SetKwComment{Comment}{/* }{ */} 
\usepackage{graphicx}
\graphicspath{{./figure/}}
\usepackage{bm}
\usepackage{amsmath,amsfonts,amssymb}
\usepackage{multirow}
\usepackage{placeins}
\usepackage{subcaption}
\usepackage{comment}
\usepackage{array}
\usepackage{stfloats}
\usepackage{booktabs}

\title{Constrained Policy Optimization for Controlled Self-Learning in Conversational AI Systems}

\author{
Mohammad Kachuee, Sungjin Lee\\
Amazon Alexa AI\\
\{kachum, sungjinl\}@amazon.com
}
\date{}

\begin{document}

\maketitle

\begin{abstract}
Recently, self-learning methods based on user satisfaction metrics and contextual bandits have shown promising results to enable consistent improvements in conversational AI systems. However, directly targeting such metrics by off-policy bandit learning objectives often increases the risk of making abrupt policy changes that break the current user experience.
In this study, we introduce a scalable framework for supporting fine-grained exploration targets for individual domains via user-defined constraints. For example, we may want to ensure fewer policy deviations in business-critical domains such as shopping, while allocating more exploration budget to domains such as music. Furthermore, we present a novel meta-gradient learning approach that is scalable and practical to address this problem. The proposed method adjusts constraint violation penalty terms adaptively through a meta objective that encourages balanced constraint satisfaction across domains.
We conduct extensive experiments using data from a real-world conversational AI on a set of realistic constraint benchmarks. Based on the experimental results, we demonstrate that the proposed approach is capable of achieving the best balance between the policy value and constraint satisfaction rate.
\end{abstract}

\section{Introduction}
Conversational AI systems such as Apple Siri, Amazon Alexa, Google Assistant, and Microsoft Cortana rely on multiple processing components for speech recognition, natural language understanding (NLU), taking proper actions, and generating a response to the user. In such a system, a skill routing block is responsible for selecting the right skill and NLU interpretation to serve the request. Skill routing is a challenging problem as thousands of skills are present in real-world conversational systems and new skills are being introduced every day. In such a scenario, gathering human annotations is very expensive and suffers from high turn-around times. Moreover, often more than one skill is capable of serving a request which makes human supervision even more challenging due to the lack of clear ground-truth assignments~\citep{sarikaya2017technology}.

Recently, self-learning methods have been proposed that leverage customer experience signals to define reward values and create a closed feedback loop~\citep{karampatziakis2019lessons}. In contrast to more traditional methods that are based on replication of rule-based systems or defect relabeling~\citep{park2020scalable}, self-learning methods continuously explore different routing alternatives and leverage user feedback to improve their decisions~\citep{kachuee2022scalable}.

Despite their scalability and efficiency, because self-learning approaches directly optimize routing decisions to achieve highest rewards, they suffer from instability issues impacting the user experience. 
Specifically, off-policy contextual bandits frequently used as the policy learning algorithm are susceptible to off-policy optimization errors, resulting in potentially breaking the current user experience due to overestimation of action values or excessive explorations~\citep{swaminathan2016off,joachims2018deep,lopez2021learning}.
Such instabilities and drastic changes in the agent's behavior not only regress user retention and trust, but also manifest as direct revenue loss for business-critical domains such as shopping.

In a production system, it is crucial to not only estimate but also control the changes of behavior a new policy introduces when compared to the current production policy. In the literature, this problem has been studied under safe bandit updates~\citep{jagerman2020safe,daulton2019thompson,amani2019linear} and budgeted bandit learning~\citep{hoffman2014correlation,guha2007multi}, usually targeting exploration budgets or encouraging a behavior resembling a baseline policy.

In the context of off-policy bandit updates, we define exploration as any change in the model behavior resulting from replacing a current production policy with a new updated policy. This definition is broad and encloses stochastic exploration actions as well as any behavior change when comparing the two consecutive policies. 
Furthermore, we consider the scenario in which samples are naturally classified into a set of domains, each representing a unique data segment. Note that in a task-oriented conversational agent, domains are typically defined based on NLU interpretation of the request (e.g. music, shopping, books).


While previous studies considered different aspects of constraining a bandit model, to the best of our knowledge the problem of controlling off-policy bandit behavior changes across subsequent model updates with a fine-grained control on budgets for different data segments (domains) remains unaddressed. This study is the first to tackle the aforementioned issues by providing a scalable and practical approach.
The main contributions of this paper are as follows:
\begin{itemize}
    \item Introducing a formulation for controlled exploration in the context of off-policy contextual bandit learning considering fine-grained control over domains based on user-defined constraints.
    \item Presenting a solution based on the primal-dual minimax constraint optimization method that is effective but requires adjusting a few hyperparameters.
    \item Proposing a novel meta gradient algorithm to balance constraint satisfaction and reward maximization that works out-of-the-box and outperforms other methods with no need for hyperparameter adjustment.
    \item Conducting extensive experiments on the skill routing problem in a real-world conversational AI agent using a set of realistic constraint benchmarks.
\end{itemize}

\section{Related Work}
\subsection{Skill Routing in Conversational AIs}
In contrast to traditional rule-based systems, model-based skill routing approaches leverage machine learning models to understand a user request and predict the best action to serve the request. The initial implementations of model-based routing started from replication of rule-based systems and improving generalization via data augmentation and relabeling ideas~\citep{li2021neural,park2020scalable}.

To improve scalability, self-learning methods have been proposed that rely on user feedback rather than human annotations to learn and improve their skill routing policies in a closed-loop. The recent work by \citet{kachuee2022scalable} is an excellent example of such approach in which model-based customer satisfaction metrics \citep{kachuee2021self} are used to define the reward function, then a stochastic mixture of replication and bandit models is used to control the exploration rate and safeguard the user experience. Nonetheless, 
such design may result in sub-optimal decisions as the bandit optimization does not consider the exploration budgets, the stochastic mixture may not be sufficiently fine-grained to protect user experience in smaller traffic segments, and deploying such architecture requires dealing with additional complexity of maintaining a separate replication model.

\subsection{Constrained Bandit Learning}
The majority of studies on controlled bandit learning consider the case of simple multi-armed stochastic bandits (i.e., without context) with practical applications in experiment design~\citep{guha2007multi} and automated machine learning~\citep{hoffman2014correlation}. \citet{hoffman2014correlation} suggested a Bayesian approach to two-phase exploration-exploitation bandit learning in which there is a pre-specified budget for exploration arm evaluations. Another aspect is to ensure safe exploration actions, which is especially useful for sensitive applications in industrial machine learning or healthcare. \citet{amani2019linear} introduced a solution in which an initial set of exploration actions is defined, then the exploration set is gradually expanded to ensure minimal unexpected behavior.

For contextual bandits, safety has been an active research topic. Safety can be defined in the action space or in terms of model updates. For example, \citet{daulton2019thompson} solves a two-metric setting in which one of the metrics, reward, is being maximized while enforcing a limit for regression on an auxiliary metric compared to a baseline status quo model. \citet{balakrishnan2018using} attempts to learn behavioral constraints by balancing between replication of the current baseline policy and making new actions that show promising rewards. In~\citep{jagerman2020safe} authors define safety in terms of user experience metrics and suggest deciding on deploying a new model based on conservative confidence bounds on the off-policy estimates of such metrics.

\section{Constrained Bandit Exploration}
\label{sec:methods}
\subsection{Problem Formulation}

We consider the general framework of off-policy contextual bandits in which a policy $\Pi$ is used to select an action $a \in A$ given the observed context vector ($\mathbf{x}$) to maximize the scalar reward ($r$) received from the environment. Here, we assume stochastic policies of the form $\Pi_{\theta}(a|\mathbf{x})$ in which a model parameterized by $\theta$ (e.g., a neural network) is used to assign action selection probabilities to each action given the context. Furthermore, we assume that each sample belongs to a domain denoted by $k \in {1 \dots M}$ that is provided as a feature in $\mathbf{x}$.

In the off-policy setting, the policy is refreshed after collecting a dataset of samples from the current policy. We adopt a definition of exploration which considers any change in the agent behavior compared to the current policy as an exploration action. 
Alternatively, we can consider replication with respect to the current policy as the rate at which the new policy makes similar decisions to the current policy when both evaluated and sampled stochastically. We define replication for $\Pi_\theta$ with respect to $\Pi_0$ based on the L1-distance of their action propensities given a context $\mathbf{x}$:
\begin{equation}
    \mathcal{R}_{\theta} (\mathbf{x}) = 1 - \frac{|\Pi_{\theta}(\mathbf{x}) - \Pi_{0}(\mathbf{x})|_1}{2} \; .
\end{equation}

In a production system, it is desirable to precisely control the rate at which the new policy replicates the current policy for each domain. This ensures robust and controlled model updates for critical domains while enabling exploration for others that may benefit from an extra exploration budget. Accordingly, we define constraints to encourage the desired behavior for samples of each domain, while learning an off-policy bandit:
\begin{equation}
\label{eq:constrained_bandit}
\begin{split}
    \arg\min_{\theta} \;\; \mathbb{E}_{\mathbf{x},a,r,k \sim \mathbb{D}} \; \; L_{\Pi_{\theta}} \;\; , \\
    s.t. \;\;\; c^{min}_{k} \leq  \mathcal{R}_{\theta} (\mathbf{x})  \leq c^{max}_{k}
\end{split}
\end{equation}
where context, action, reward, and domain ($\mathbf{x},a,r,k$) are sampled from a dataset collected from the current policy.
In \eqref{eq:constrained_bandit}, we use $c^{min}_{k}$ and $c^{max}_{k}$ to indicate user-defined replication constraints for domain $k$.

$L_{\Pi_{\theta}}$ can be any differentiable off-policy bandit learning objective, for simplicity of discussion, we consider the vanilla inverse propensity scoring (IPS) objective:
\begin{equation}
\label{eq:ips}
    L_{\Pi_{\theta}}(\mathbf{x},a,r) = -r  \frac{\Pi_\theta(a|\mathbf{x})}{\Pi_0(a|\mathbf{x})} \; \;,
\end{equation}
where $\Pi_0$ is the current policy and $r$ is the observed reward for taking action $a$ collected in the dataset.

A common approach to optimize constrained problems such as \eqref{eq:constrained_bandit} is to use the penalty method, translating constraints into penalty terms that encourage constraint satisfaction:
\begin{multline}
\label{eq:constrained_bandit_penalty}
    \arg\min_{\theta} \;\; \mathbb{E}_{\mathbf{x},a,r,k \sim \mathbb{D}} \; \; [L_{\Pi_{\theta}}(\mathbf{x},a,r) \; + \\
     e^{\mathbf{u}_k} \max (0, c^{min}_{k} - \mathcal{R}_{\theta}(\mathbf{x})) \; + \; \\ e^{\mathbf{v}_k} \max (0,   \mathcal{R}_{\theta}(\mathbf{x}) - c^{max}_{k})] \;\;.
\end{multline}

Here, penalty terms are always non-negative and increase if the new policy assigns action probabilities that deviate from the current policy outside the desired boundary. $\mathbf{u} \in R^{M}$ and $\mathbf{v} \in R^{M}$ are variables that adjust the weight of each constraint violation term. The exponentiation improves the sensitivity to these parameters and ensures having non-negative penalty terms. For \eqref{eq:constrained_bandit_penalty} to actually solve the original constrained problem of \eqref{eq:constrained_bandit}, proper values for $\mathbf{u}$ and $\mathbf{v}$ need to be used that enable the best balance between constraint satisfaction and the policy value. In the constrained optimization literature, various methods have been suggested to solve this form of problem. In this paper, to solve this problem,  we use the primal-dual minimax method suggested by \citet{nandwani2019primal} (Section~\ref{sec:minimax}) as well as a novel meta-learning method (Section~\ref{sec:meta}).


\subsection{Minimax Primal-Dual Method}
\label{sec:minimax}
\citet{nandwani2019primal} suggested a formulation of the augmented Lagrangian method that supports inequality constraints. They solve the dual problem which is optimizing the dual maximin problem to improve the scalability:
\begin{multline}
\label{eq:constrained_bandit_penalty_minimax}
    \min_{\theta} \;\; \max_{\mathbf{u} , \mathbf{v}} \;\; \mathbb{E}_{\mathbf{x},a,r,k \sim \mathbb{D}} \; \; [L_{\Pi_{\theta}}(\mathbf{x},a,r) \; + \\
     e^{\mathbf{u}_k} \max (0, c^{min}_{k} - \mathcal{R}_{\theta}(\mathbf{x})) + \\ e^{\mathbf{v}_k} \max (0,   \mathcal{R}_{\theta}(\mathbf{x}) - c^{max}_{k})] \;\;.
\end{multline}
Algorithm~\ref{alg:minimax} shows an outline of the policy training using the minimax method. This method has four hyperparameters controlling the max player optimization via adjusting the update frequency, learning rate, and decay factors.

\begin{algorithm}[ht]
\caption{Minimax constrained bandit optimization algorithm}
\label{alg:minimax}

\DontPrintSemicolon
\SetKwInOut{Input}{input}
\SetKwInOut{Output}{output}

\Input{$\mathbb{D}$ (dataset), $\eta$ (max learning rate), $\gamma$ (max learning rate decay), $\tau$ (max update frequency), $\xi$ (max update frequency decay)}
$\mathbf{u} , \mathbf{v}, t \gets 0$ \;
Initialize($\Pi_{\theta}$) \;
\For{$\mathbf{x}, a, r, k \sim {D}$}{
\Comment*[l]{loss function of \eqref{eq:constrained_bandit_penalty_minimax}}
  $L \leftarrow Loss(\mathbf{x}, a, r, k, \theta, \mathbf{u} , 
  \mathbf{v})$ \;
  \If{$t \% \tau$ is 0}{ 
      \Comment*[l]{gradient ascent, max player}
      $\mathbf{u} \leftarrow \mathbf{u} + \eta \nabla_{\mathbf{u}}L $ \;
      $\mathbf{v} \leftarrow \mathbf{v} + \eta \nabla_{\mathbf{v}}L $ \;
      \Comment*[l]{lr/update decay}
      $\eta \leftarrow \gamma \times \eta $ \;
      $\tau \leftarrow \xi \times \tau $ \;
  }
  \Comment*[l]{optimize $\Pi_{\theta}$, min player}
  $\theta \leftarrow f(\theta, \nabla_{\theta}L)$ \;
  \Comment*[l]{increment counter}
  $t \leftarrow t + 1$ \;
}
\end{algorithm}

Intuitively, the min player is trying to update the policy parameters while the max player is increasingly penalizing it for any constraint violation. A stable point of this algorithm would be to gradually reduce the max player update rate as the min player is getting better at satisfying the constraints, eventually satisfying all constraints resulting in a zero loss for the max player due to the zero hinge penalty terms.

\subsection{Meta Gradient Method}
\label{sec:meta}

Theoretically, the primal-dual minimax method is capable of achieving Pareto optimal solutions \citep{jin2019minmax,nandwani2019primal}. However, in practice, it is infeasible to train for an infinite number of iterations, and therefore approximate inner optimization loops are being used. To find the right balance between constraint satisfaction and policy improvement for the minimax algorithm, it is necessary to carefully adjust multiple hyperparameters.
Note that an extensive hyperparameter search is undesirable in many real-world large-scale scenarios that require frequent model updates as it entails not only significant compute costs associated with the search but also increases the turn-around time to deploy refreshed models.
To mitigate this issue, we suggest a meta-gradient optimization idea that adapts $\mathbf{u}$ and $\mathbf{v}$ based on a meta objective within the training process.

Specifically, we define the following meta objective:
\begin{multline}
\label{eq:meta}
    L_{meta} = \mathbb{E}_{\mathbf{x},a,r,k \sim \mathbb{D}} \; \; 
    (1-\lambda) L_{\Pi_{\theta}}(\mathbf{x},a,r) \; + \\
    \lambda \frac{\max (0, c^{min}_{k} - \mathcal{R}_{\theta}(\mathbf{x})) + \max (0,   \mathcal{R}_{\theta}(\mathbf{x}) - c^{max}_{k})} {p(k)}
\end{multline}

where $\lambda$ is a hyperparameter to balance between the bandit objective and the constraint penalty terms. The second term is the macro average of constraint violation terms, in which $p(k)$ is the prior probability of samples belonging to domain $k$ that can be easily pre-computed for a large batch of samples. 

Note that \eqref{eq:meta} is not directly dependent on $\mathbf{u}$ and $\mathbf{v}$, instead we rely on online cross-validation \citep{sutton1992adapting,xu2018meta} to update these variables. We define an inner objective the same as the min optimization problem of~\eqref{eq:constrained_bandit_penalty_minimax}, do a differentiable optimization step, evaluate the meta objective on another batch of data, then update $\mathbf{u}$ and $\mathbf{v}$ by taking the derivative of the meta objective through the inner optimization trace.

Algorithm~\ref{alg:meta} presents an outline of the meta gradient optimization method. Due to practical issues of dealing with high-order gradients, we only consider the immediate impact of a single inner loop update on the meta objective. We found that discarding the vanilla gradient descent used for the inner optimization and using a more advanced optimizer (e.g., Adam) to update $\Pi_{\theta}$ works best. Regarding the $\lambda$ hyperparameter, we found that simply setting $\lambda=1$ works well in practice. It effectively means that the meta-gradient solution does not require any hyperparameter adjustments (experimental evidence presented in Section~\ref{sec:results}).

\begin{algorithm}[ht]
\caption{Meta gradient constrained bandit optimization algorithm}
\label{alg:meta}

\DontPrintSemicolon
\SetKwInOut{Input}{input}
\SetKwInOut{Output}{output}

\Input{$\mathbb{D}$ (dataset), $\eta$ (learning rate), $\lambda$ (penalty weight)}
$\mathbf{u} , \mathbf{v} \gets 0$ \;
Initialize($\Pi_{\theta}$) \;
\For{$\mathbf{x}, a, r, k \sim \mathbb{D}$ and $\mathbf{x'}, a', r', k' \sim \mathbb{D}$}{
    \Comment*[l]{clone parameters}
    $\theta' \leftarrow clone(\theta)$ \;
    \Comment*[l]{inner loss with $\theta'$}
    $L_{inner} \leftarrow Loss_{inner}(\mathbf{x}, a, r, k, \theta', \mathbf{u} , \mathbf{v})$ 
    \Comment*[l]{gradient descent on cloned model}
    $\theta' \leftarrow \theta' - \eta \nabla_{\theta'}L_{inner} $ \;
    \Comment*[l]{compute meta loss}
    $L_{meta} \leftarrow Loss_{meta}(\mathbf{x'}, a', r', k', \theta', \lambda)$ \;
    \Comment*[l]{diff. through inner update}
    Compute $\nabla_{\mathbf{u}}L_{meta}$ and $\nabla_{\mathbf{v}}L_{meta}$ \;
    \Comment*[l]{optimize $\mathbf{u,v}$ using any optimizer}
    $\mathbf{u} \leftarrow f(\mathbf{u},\nabla_{\mathbf{u}}L_{meta}) $ \;
    $\mathbf{v} \leftarrow f(\mathbf{v},\nabla_{\mathbf{v}}L_{meta}) $ \;
    \Comment*[l]{inner loss with $\theta$}
    $L \leftarrow Loss_{inner}(\mathbf{x}, a, r, k, \theta, \mathbf{u} , \mathbf{v})$ \;
    \Comment*[l]{optimize $\Pi_{\theta}$ using any optimizer}
    $\theta \leftarrow f(\theta,\nabla_{\theta}L)$ \;
}
\end{algorithm}

Intuitively, at each training iteration, the inner objective naturally minimizes the bandit loss that is penalized by constraint violation terms proportional to the current $\mathbf{u}$/$\mathbf{v}$. Then, the meta objective computes a validation loss that measures the impact of the inner policy update and $\mathbf{u}$/$\mathbf{v}$ on the macro-averaged constraint violations. Finally, by computing the meta-gradient of the meta objective through the inner optimization loop, $\mathbf{u}$ and $\mathbf{v}$ are updated to better encourage the constraint satisfaction for the next policy update iteration. 
Thanks to the online cross-validation update for $\mathbf{u}$ and $\mathbf{v}$, the meta-gradient method adjusts the penalty weights such that their value does not unnecessarily keep increasing when it does not result in further improvements to the constraint satisfaction.

\section{Experiments}
\subsection{Setup
}
We conduct experiments on a bandit agent for the problem of skill routing in commercial conversational systems such as Apple Siri, Amazon Alexa, Google Assistant, and Microsoft Cortana. In these systems, skill routing is a key component that takes the user's utterance as well as signals from automated speech recognition (ASR) and natural language understanding (NLU), then it decides which skill and NLU interpretation to be used to serve the request~\citep{sarikaya2017technology}.

In a commercial dialogue agent making controlled policy updates is crucial because any change in the skill routing policy directly impacts the user experience. Making abrupt policy changes may negatively impact user retention and in certain business-critical domains may result in loss of revenue. In such scenario, constraints can be defined based on the NLU domain to ensure policy robustness for business-critical domains such as shopping, while potentially exploring others such as entertainment more aggressively.

\begin{figure}[t]
    \centering
        \includegraphics[width=0.8\linewidth]{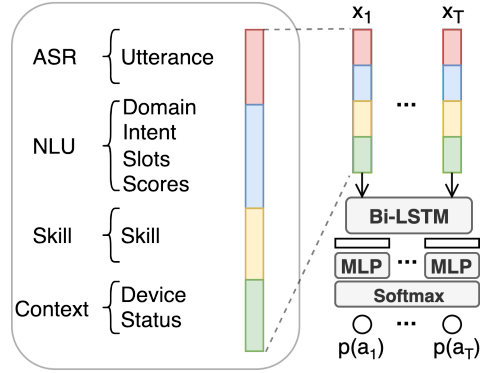}
        \caption{An overview of the network architecture: a set of hypothesis are encoded as vectors and fed to a bi-directional LSTM which is followed by a shared MLP and a softmax layer to normalize the candidate selection probabilities.}
        \label{fig:model_arch}
\end{figure}

Figure~\ref{fig:model_arch} shows an overview of the model architecture used in our experiments. Input to the model is a set of routing candidates i.e., a combination of embedded ASR, NLU, and context vectors as well as skill embeddings. The output is the softmax-normalized propensity of selecting each candidate to handle the user request.
The final model has about 12M trainable parameters consisting of a language model to encode utterance, embeddings for contextual signals, and fully-connected layers.
As our architecture closely follows the design choices from \citet{kachuee2022scalable}, we refer interested readers to that paper for details.

To train and evaluate our models, we use logged policy actions from a current production policy. The observed reward is based on a curated function of user satisfaction metrics~\citep{kachuee2021self}. Our dataset consists of about 40M samples divided into 85\% training, 10\% validation, and 5\% test sets covering 27 domains that are imbalanced in the number of samples. All data used in this work was deidentified to comply with our customer privacy guidelines.

\subsection{Benchmarks}
\begin{figure}[t]
    \centering
        \includegraphics[width=0.95\linewidth]{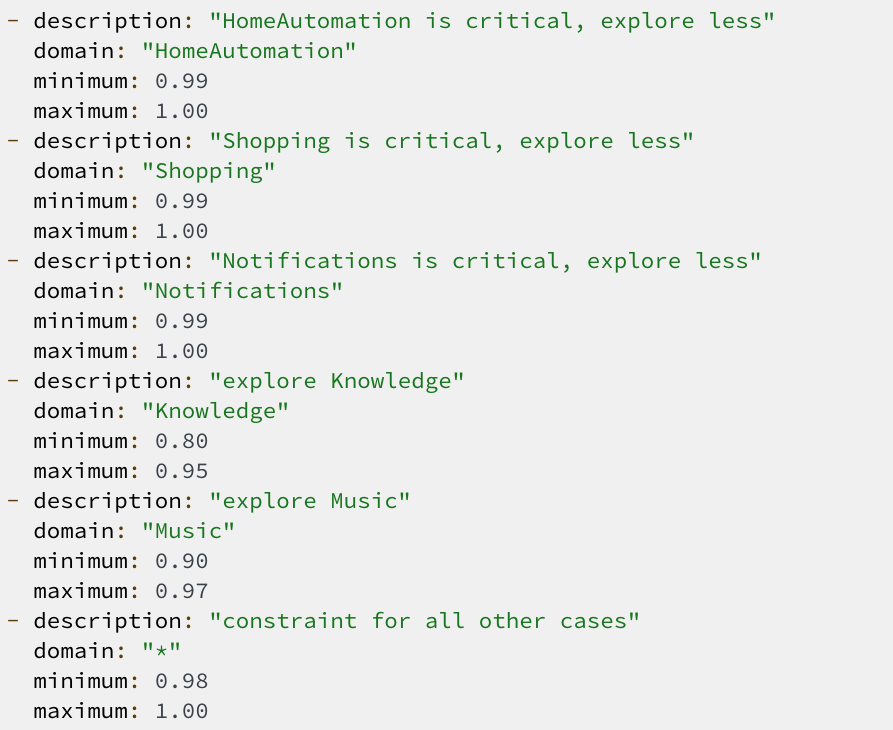}
        \caption{The constraint configuration list for the \texttt{exploration} benchmark.}
        \label{fig:constraint_example}
\end{figure}

In our experiments, we use three different benchmarks for the constraint settings: \texttt{global}, \texttt{critical}, and \texttt{exploration}. The \texttt{global} benchmark aims to constrain the new policy to be within an exploration limit for all domains. In addition to the global constraint,  \texttt{critical} assert stronger limits for a set of critical domains defined based on the expert knowledge. The \texttt{exploration} benchmark extends the critical benchmark by adding constraints to encourage exploration for domains that may benefit from additional exploration. Each benchmark is a list of constraints consisting of a short description, applicable domain, and the desired replication range. Figure~\ref{fig:constraint_example} shows the \texttt{exploration} benchmark as an example. We provide the exact constraint configurations in the appendix.

\subsection{Baselines and Metrics}
As the first baseline, we consider the vanilla IPS objective which disregards the constraints. Additionally, we build on the IPS baseline to consider the constraints using constraint optimization approaches: quadratic (uniform constant penalty weight), minimax (Algorithm~\ref{alg:minimax}), and meta-gradient (Algorithm~\ref{alg:meta}). Unless expressed otherwise, we use Adam optimizer with the default configuration~\citep{kingma2014adam} (denoted by $f$ in Section~\ref{sec:methods}).

Regarding hyperparameters, for the penalty weight of the quadratic method we use values from $\{0.1, 1, 10, 100, 1000\}$. For the minimax method (Algorithm~\ref{alg:minimax}), we found that setting $\tau$ and $\xi$ to one while adjusting $\eta$ and $\gamma$ presents very similar results to adjusting all four hyperparameters. Consequently, we use a grid search of  $\eta \in \{1, 0.1, 0.01\}$ and $\gamma \in \{1, 0.999, 0.995 \}$ to find the best settings for each benchmark. For the meta-gradient method (Algorithm~\ref{alg:meta}), we found that simply using $\lambda$ equal to one in the meta objective (i.e., meta objective only focusing on the constraints) outperforms other works (see evidence presented in Section~\ref{sec:results}). As a result, it does not require adjusting any hyperparameter and the same setting is used across all benchmarks. We provide the final hyperparameters used for each benchmark and method in the appendix.

Regarding the evaluation metrics, we use the expected off-policy reward as well as the rate of change in constraint violations averaged over all samples (i.e., micro-averaged) and individual domains (i.e., macro-averaged). To comply with our privacy and business guidelines, in all instances, we only report relative and normalized results which do not directly represent the actual scales or metric values.

We train each model until convergence or reaching 32 epochs and take the best performing model based on the macro-averaged violation rate measured on the validation set. Each experiment was run four times using different random seeds for data sampling and weight initialization to report the mean and standard deviation of each result. We used a cluster of 32 NVIDIA V100 GPUs to process a mini-batch size of 32K samples. Each individual run took between 4 to 24 hours.

\subsection{Results}
\label{sec:results}

\subsubsection{Offline Experimentation}
Table~\ref{tab:benckmark_results} shows a comparison of results for the IPS, quadratic, minimax, and meta-gradient methods on all benchmarks. For each case, we report the expected reward and the percentage of reduction in the rate of violations compared to the simple IPS objective. The meta-gradient approach consistently shows the best results across all benchmarks. The simple quadratic method behaves very competitively to minimax, except for the \texttt{explore} benchmark which requires a more fine-grained control on multiple constraints (see Figure~\ref{fig:constraint_example}). The meta-gradient method, while having the highest reduction in constraints violations, also has very competitive performance in terms of the reward metric.

\begin{table*}[t]
\centering
\resizebox{\textwidth}{!}{
\renewcommand{\arraystretch}{1.3}
\begin{tabular}{l|ccc|ccc|ccc}
\toprule
& \multicolumn{9}{c}{\textbf{Benchmark}} \\
& \multicolumn{3}{c|}{\texttt{global}} & \multicolumn{3}{c|}{\texttt{critical}} & \multicolumn{3}{c}{\texttt{explore}} \\
\textbf{Method} & reward & \multicolumn{2}{c|}{violation reduction} & reward & \multicolumn{2}{c|}{violation reduction} & reward & \multicolumn{2}{c}{violation reduction} \\
 & $(\%)$ & macro (\%) & micro (\%) & $(\%)$ & macro (\%) & micro (\%) & $(\%)$ & macro (\%) & micro (\%) \\  
 \hline
IPS & 
89.45{\footnotesize$\pm${0.01}} & 0 & 0 
& 89.45{\footnotesize$\pm${0.01}} & 0 & 0
& 89.45{\footnotesize$\pm${0.01}} & 0 & 0\\
Quadratic & 
88.95{\footnotesize$\pm${0.01}} & 63.67{\footnotesize$\pm${0.46}} & 63.67{\footnotesize$\pm${0.46}}
& 88.94{\footnotesize$\pm${0.01}} & 50.13{\footnotesize$\pm${0.90}} & 69.29{\footnotesize$\pm${0.67}}
& 88.36{\footnotesize$\pm${0.04}} & 28.37{\footnotesize$\pm${4.62}} & 65.24{\footnotesize$\pm${2.30}}\\
Minimax & 88.91{\footnotesize$\pm${0.01}} & 63.28{\footnotesize$\pm${0.08}} & 63.28{\footnotesize$\pm${0.08}}
& 88.93{\footnotesize$\pm${0.01}} & 37.88{\footnotesize$\pm${0.49}} & 62.51{\footnotesize$\pm${0.21}}
& 88.11{\footnotesize$\pm${0.01}} & 61.51{\footnotesize$\pm${0.59}} & 81.50{\footnotesize$\pm${0.24}}\\
MetaGrad & 88.94{\footnotesize$\pm${0.01}} & \textbf{75.91}{\footnotesize$\pm${0.49}} & \textbf{75.91}{\footnotesize$\pm${0.49}}
& 88.94{\footnotesize$\pm${0.01}} & \textbf{60.63}{\footnotesize$\pm${0.95}} & \textbf{79.69}{\footnotesize$\pm${0.85}}
& 88.41{\footnotesize$\pm${0.01}} & \textbf{78.23}{\footnotesize$\pm${0.17}} & \textbf{89.95}{\footnotesize$\pm${0.20}}\\
\hline
\end{tabular}
}
\caption{A comparison of the baseline IPS method with the quadratic, minimax, and meta-gradient constrained optimization methods on different benchmarks. We report the normalized percentage of reduction in the number of constraint violations compared to the IPS method.} 
\label{tab:benckmark_results}
\end{table*}

\subsubsection{Online Experimentation}
We conducted an A/B experiment to compare the proposed method with the stochastic gating method of~\citet{kachuee2022scalable} for robust self-learning (indicated by RPDR in the table). We conducted our A/B in two phases, deploying and comparing each approach to a baseline skill routing production system. Each phase took one week and consisted of traffic from about 6M customers (3M control and 3M treatment). For the RPDR method, we used a target replication rate of $99\%$ for each domain. The meta-gradient model was trained with the \texttt{global} benchmark, constraining to a similar $99\%$ replication. For both RPDR and MetaGrad models, we used the same training set which was collected from the control model behavior and followed the same model architecture.

Table~\ref{tab:ab_results} presents the results of the A/B experiment. For each method we report the percentage of changes in the achieved reward compared to the control model. For violation reduction, we report the percentage of reduction for MetaGrad compared to the RPDR method. For the replication metric, we simply report the percentage of time that each policy makes actions that replicate the control model decision. As we can see from the results, MetaGrad approach not only shows more stable behavior by better constraint satisfaction and replication rates, but it also achieves statistically significant improvements in the reward value.

\begin{table}[t]
\centering
\resizebox{\linewidth}{!}{
\renewcommand{\arraystretch}{1.3}
\begin{tabular}{l|ccc}
\toprule
\textbf{Method} & reward & Violation Reduction & Replication \\
 & (\%) & (\%) & (\%) \\  
 \hline
RPDR & -0.01 {\footnotesize{(p>0.05)}} & 0 & 98.13 \\
MetaGrad & +0.19 {\footnotesize{(p<0.05)}} & 38.05 & 99.11 \\
\hline
\end{tabular}
}
\caption{Comparison of the proposed method (MetaGrad) and the robust self-learning method by ~\citet{kachuee2022scalable} (RPDR) using an online A/B experiment. We report: percentage of change in the reward compared to a control model, violation reduction for the MetaGrad normalized by the RPDR result, and percentage of replication compared to the control policy actions.}
\label{tab:ab_results}
\end{table}

\subsubsection{Impact of Hyperparameters}
\begin{figure*}[h]
    \centering
    \begin{subfigure}[b]{0.3\linewidth}
        \centering
        \includegraphics[width=\columnwidth]{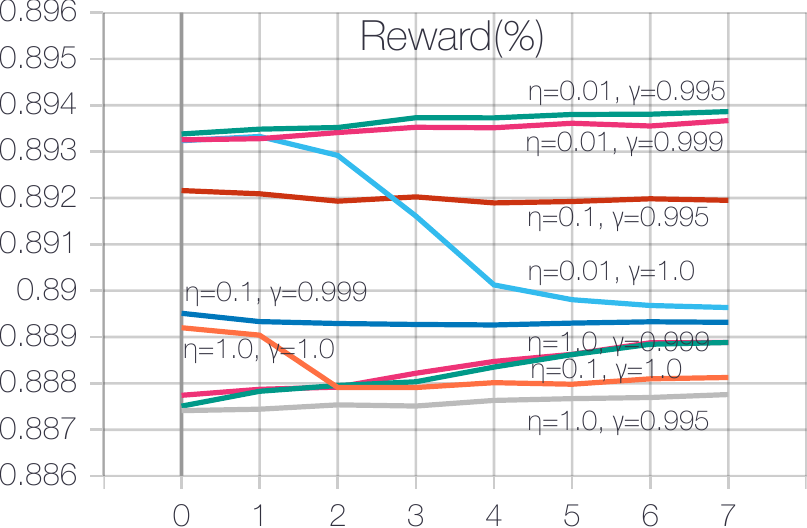}
        \caption{}
    \end{subfigure}
    ~
    \begin{subfigure}[b]{0.3\linewidth}
        \centering
        \includegraphics[width=\columnwidth]{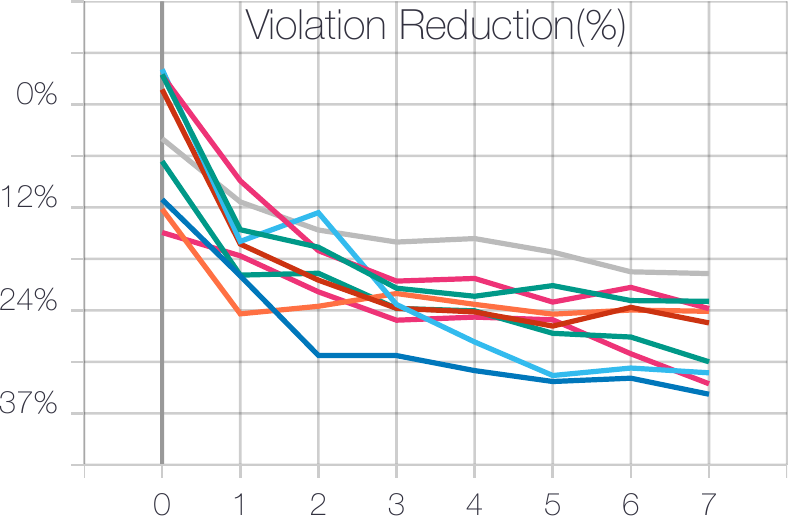}
        \caption{}
    \end{subfigure}
    \\
    \begin{subfigure}[b]{0.3\linewidth}
        \centering
        \includegraphics[width=\columnwidth]{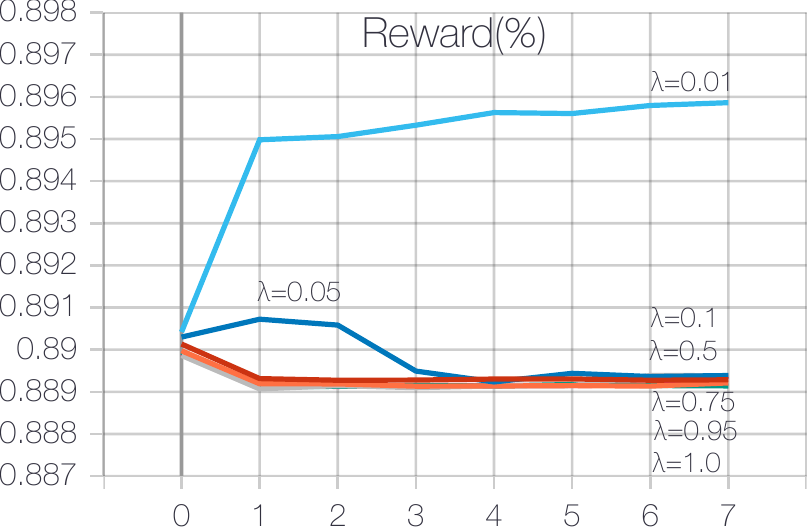}
        \caption{}
    \end{subfigure}
    ~
    \begin{subfigure}[b]{0.3\linewidth}
        \centering
        \includegraphics[width=\columnwidth]{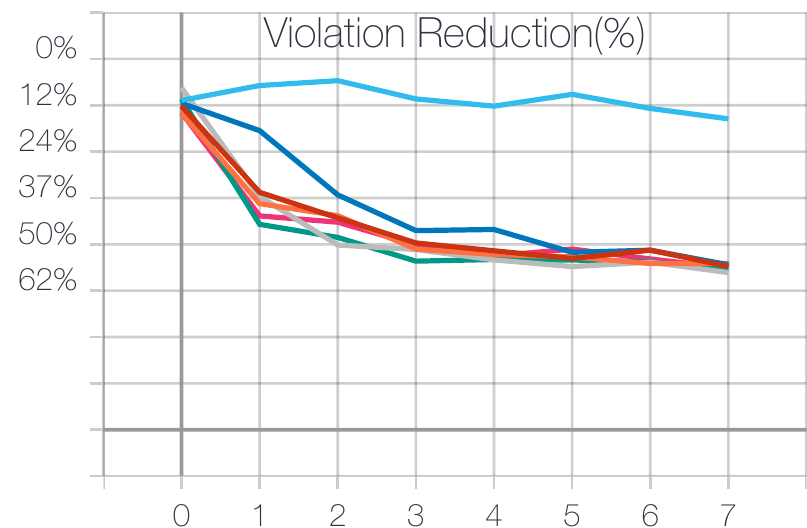}
        \caption{}
    \end{subfigure}
    \caption{Comparing the hyper-parameter sensitivity for the minimax and meta-gradient methods on the \texttt{critical} benchmark. For the minimax method: (a) reward and (b) macro violation reduction wrt. different $\eta$ and $\gamma$ settings. For the meta-gradient method: (c) reward and (d) macro violation reduction wrt. different $\lambda$ settings.}
    \label{fig:hyper_param_impact}
\end{figure*}

To study the impact of hyperparameters, we conducted an experiment using the \texttt{critical} benchmark by training minimax and meta-gradient based models using different hyperparameter values. Specifically, we train minimax models (Algorithm~\ref{alg:minimax}) using $\eta \in \{1.0, 0.1, 0.01\}$ and $\gamma \in \{1.0, 0.999, 0.995 \}$. For the meta-gradient method (Algorithm~\ref{alg:meta}), we use  $\lambda \in \{0.01, 0.05, 0.1, 0.5, 0.75, 0.95, 1.0 \}$. Figure~\ref{fig:hyper_param_impact} shows the results of such experiment. Based on this experiment, the minimax approach shows a much higher sensitivity to its two hyperparameters, showing a significant impact on both the reward and violation reduction metrics. However, the meta-gradient method shows much less sensitivity to the $\lambda$ hyperparameter. We found that simply setting $\lambda=1$ works very well in practice. It can be very desirable for real-world large-scale settings such as conversational systems which require frequent model updates as new features are on-boarded every day, and having a dependency on an extensive hyperparameter search is very costly, if not impractical.

\subsubsection{Analysis of Penalty Weights}
\begin{figure}[h]
    \centering
        \includegraphics[width=0.75\linewidth]{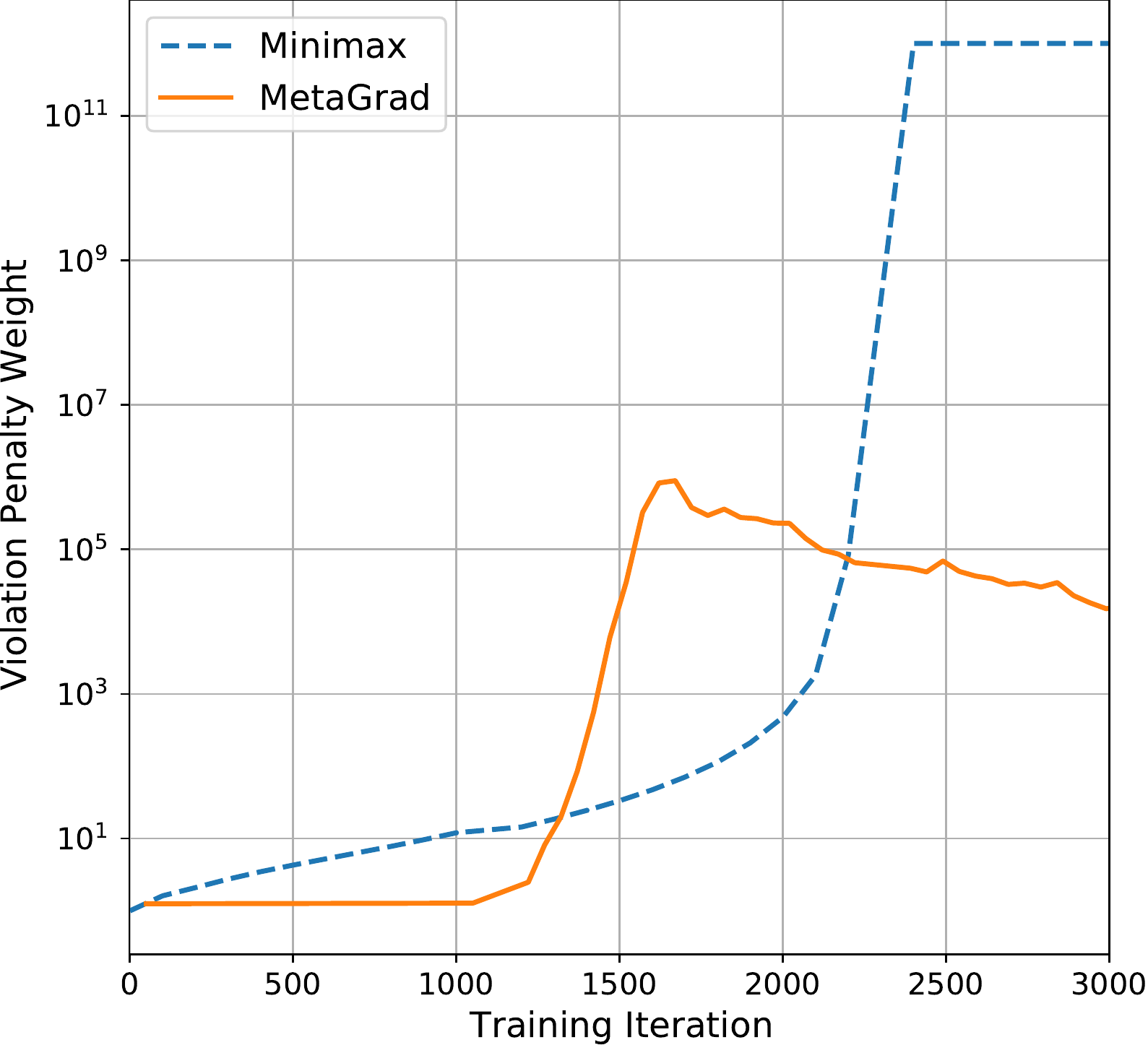}
        \caption{The constraint penalty weight values for the first 3,000 iterations of training using the \texttt{global} benchmark.}
        \label{fig:global_penalty}
\end{figure}

To dive deeper into the reason behind the better performance for the meta-gradient algorithm compared to the minimax approach, we investigated the constraint penalty weight value for the first 3,000 iterations of training using the \texttt{global} benchmark. From Figure~\ref{fig:global_penalty}, we can see the minimax method is monotonically increasing the penalty weight with each iteration which is a behavior consistent with the gradient ascent update rule in Algorithm~\ref{alg:minimax}. In other words, as long as there are any constraint violations, minimax will keep increasing the penalty, which in our opinion is the reason for high sensitivity to the hyperparameters. On the other hand, the meta-gradient approach is using a validation signal to dynamically adjust the penalty weight. Consequently, it may keep the penalty term near zero for an initial phase, rapidly increase it, then decay when violations are reduced and getting a higher reward is preferred.

\section{Conclusion}
This work studied the problem of controlled exploration for off-policy contextual bandit learning to control the policy update robustness in self-learning skill routing systems. We presented a constraint optimization formulation that enables a human expert to define the boundary of the desired exploration rate for individual domains. We proposed a scalable and practical solution based on meta-gradient learning which provides the highest constraint satisfaction rates without any need for an extensive hyperparameter adjustment. Finally, we conducted experiments using data from a real-world conversation system for the skill routing problem on a set of different realistic constraint benchmarks. Based on the experimental results, we believe that the suggested controlled bandit learning approach is very promising for application in real-world bandits for use-cases such as conversational AIs in which frequent but controlled policy updates are of paramount importance.

\section{Limitations}
While we conducted extensive experiments and demonstrated the effectiveness of the suggested approach for controlled bandit learning in the context of the skill routing problem, there are multiple directions of improvement for future studies. 
We believe one of the limitations of the suggested constrained optimization framework is that it relies on expert-defined conditions on an arbitrary segmentation of samples. It entails the need for human intervention and manual constraint definition/optimization which can be challenging. Another limitation we faced was during our experiments which showed additional compute overhead of between 2 to 3 times for different constrained optimization methods due to additional optimization objectives, inner loops, and backward passes.

\bibliography{refs}

\clearpage
\appendix

\section{Appendix}
\subsection{Constraint Benchmarks}
\begin{figure}[h]
    \centering
    \begin{subfigure}[t]{0.95\columnwidth}
        \centering
        \includegraphics[width=\columnwidth]{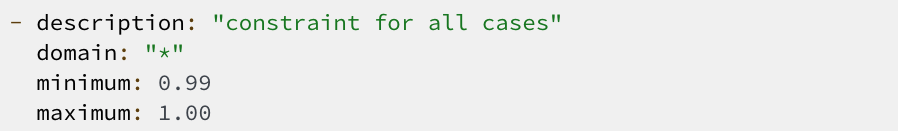}
        \caption{\texttt{global} benchmark}
    \end{subfigure}
    \\
    \centering
    \begin{subfigure}[t]{0.95\columnwidth}
        \centering
        \includegraphics[width=\columnwidth]{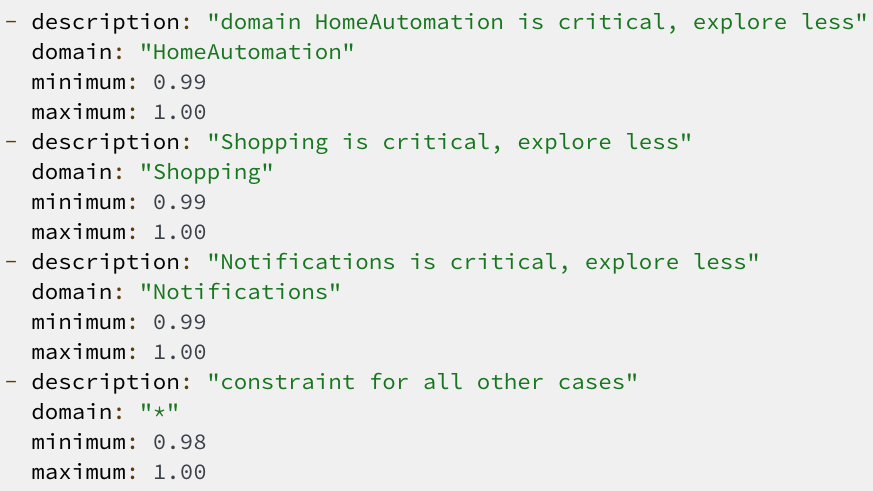}
        \caption{\texttt{critical} benchmark}
    \end{subfigure}
    \\
    \begin{subfigure}[t]{0.95\columnwidth}
        \centering
        \includegraphics[width=\columnwidth]{figures/benchmark_explore_2.png}
        \caption{\texttt{explore} benchmark}
    \end{subfigure}
        \caption{The constraint benchmarks used in this paper: (a) \texttt{global}, (b) \texttt{critical}, and (c) \texttt{explore}.}
        \label{fig:constraints}
\end{figure}

Figure~\ref{fig:constraints} presents the definition of constraint benchmarks used in this paper: \texttt{global}, \texttt{critical}, and \texttt{explore}. The \texttt{global} benchmark sets a general minimum replication rate for all domains. The \texttt{critical} benchmark defines a tighter minimum replication rate for three business-critical domains (home automation, shopping, and notifications) and a more relaxed default case for all other domains. In the \texttt{explore} benchmark, we extend the \texttt{critical} benchmark to include exploration encouragement for the knowledge and music domains.

\subsection{Selected Hyperparameters}
Table~\ref{tab:selected_hyperparameters} shows the final selected hyperparameters for each benchmark and method. The definition of each hyper-parameter is presented in Algorithm 1 and 2.

\begin{table}[t]
\centering
\resizebox{\columnwidth}{!}{
\renewcommand{\arraystretch}{1.3}
\begin{tabular}{lcccc}
\toprule
 & & \multicolumn{3}{c}{\textbf{Benchmark}} \\
\textbf{Method} & & {\texttt{global}} & {\texttt{critical}} & {\texttt{explore}} \\
\hline
Quadratic & $w$ & 10 & 1000 & 1000 \\
\hline
\multirow{2}{*}{Minimax} & $\eta$ & 0.1 & 0.1 & 1 \\
& $\gamma$ & 1 & 0.999 & 1 \\
\hline
Meta-Grad & $\lambda$ & 1 & 1 & 1 \\
\hline
\end{tabular}
}
\caption{The selected hyperparameters for each benchmark and method.} 
\label{tab:selected_hyperparameters}
\end{table}


\end{document}